\documentclass[11pt]{article}

\usepackage[final]{acl}

\usepackage{times}
\usepackage{latexsym}

\usepackage[T1]{fontenc}
\usepackage[utf8]{inputenc}

\usepackage{microtype}

\usepackage{inconsolata}

\usepackage{graphicx}

\usepackage[utf8]{inputenc}
\usepackage[T1]{fontenc}
\usepackage{algorithm}
\usepackage{algorithmic}
\usepackage{graphicx}
\usepackage{booktabs}
\usepackage{multirow}
\usepackage{amsmath}
\usepackage{amsfonts}
\usepackage{enumitem}
\usepackage{url} 
\usepackage{xspace}
\usepackage[table]{xcolor}
\usepackage{hyperref}
\usepackage{epigraph}
\usepackage{amsthm}
\usepackage{amssymb}
\usepackage{arydshln}
\usepackage{bbm}
\usepackage{wrapfig}
\usepackage{subcaption}

\theoremstyle{plain}

\theoremstyle{definition}

\theoremstyle{remark}

\usepackage[most]{tcolorbox}
\usepackage{xcolor}

\definecolor{langdark}{rgb}{0, 0, 0}
\definecolor{langlightblue}{rgb}{0.3, 0.65, 1}
\definecolor{langblue}{rgb}{0, 0.4, 0.8}
\definecolor{langmildblue}{rgb}{0.0, 0.45, 0.73}
\definecolor{langdarkblue}{rgb}{0.0, 0.0, 0.61}
\definecolor{langred}{rgb}{0.81, 0.09, 0.13}
\definecolor{langgreen}{rgb}{0.18, 0.55, 0.34}
\definecolor{langdarkgreen}{rgb}{0.0, 0.45, 0.38}
\definecolor{bingpink}{rgb}{1.0, 0.41, 0.71}
\definecolor{lightgray}{gray}{0.92}
\hypersetup{
    colorlinks=true,
    linkcolor=langdark,
    citecolor=langgreen,
    urlcolor=langdarkgreen,
}

\newcommand{\gain}[1]{\textsubscript{\textbf{\textit{\textcolor{langgreen}{+#1}}}}}
\newcommand{\loss}[1]{\textsubscript{\textbf{\textit{\textcolor{langred}{-#1}}}}}

\makeatletter
\AtBeginDocument{%
  \let\oldref\ref
  \renewcommand{\ref}[1]{%
    \hyperref[{#1}]{\underline{\oldref{#1}}}%
  }%
}
\makeatother

\usepackage{titletoc}
\newcommand\DoToC{%
  \startcontents
  \printcontents{}{1}{\textbf{\large Contents of Appendix}\vskip3pt\hrule\vskip5pt}
  \vskip3pt\hrule\vskip5pt
}
\title{RAG or Learning? Understanding the Limits of LLM Adaptation under Continuous Knowledge Drift in the Real World}

\newcommand{\method}{\textit{Chronos}\xspace}

\author{
Hanbing Liu$^1$\thanks{Equal contribution. Author order may be swapped.}\thanks{Affiliated with the Shenzhen Key Laboratory of Ubiquitous Data Enabling, Tsinghua Shenzhen International Graduate School, Tsinghua University.}
\quad
Lang Cao$^2$\footnotemark[1]
\quad
Yang Li$^3$\thanks{Corresponding author.}
\\
$^1$Tsinghua University
\quad
$^2$University of Illinois Urbana-Champaign \\
\quad
$^3$School of Artificial Intelligence, Chinese University of Hong Kong (Shenzhen)
\\
\texttt{liuhb24@mails.tsinghua.edu.cn \quad langcao2@illinois.edu \quad yangl@cuhk.edu.cn}
}

\begin{document}
\maketitle

% Keywords: Continuous Knowledge Drift, Model Adaptation, Large Language Models
% TLDR: This work introduces a new benchmark for evaluating LLMs under continuous knowledge drift and propose Chronos, a time-aware retrieval framework that enables temporally consistent reasoning over evolving real-world information without additional training.

\begin{abstract}
Large language models (LLMs) acquire most of their knowledge during pretraining, which ties them to a fixed snapshot of the world and makes adaptation to continuously evolving knowledge challenging. As facts, entities, and events change over time, models may experience \emph{continuous knowledge drift}, resulting not only in outdated predictions but also in temporally inconsistent reasoning. Although existing approaches, such as continual finetuning, knowledge editing, and retrieval-augmented generation (RAG), aim to update or supplement model knowledge, they are rarely evaluated in settings that reflect chronological, evolving, and real-world knowledge evolution. In this work, we introduce a new benchmark of real-world dynamic events, constructed from time-stamped evidence that captures how knowledge evolves over time, which enables systematic evaluation of model adaptation under continuous knowledge drift. The benchmark reveals that most existing methods, including vanilla RAG and several learning-based approaches, struggle under this setting, exposing critical limitations such as catastrophic forgetting and temporal inconsistency. To mitigate these limitations, we propose a time-aware retrieval baseline, \method, which progressively organizes retrieved evidence into an Event Evolution Graph to enable more temporally consistent understanding in LLMs without additional training. Overall, this work provides a foundation for analyzing and advancing LLM adaptation to continuous knowledge drift in realistic settings. Code is available at \url{https://github.com/hbing-l/chronos}.
\end{abstract}
% \footnote{Code is available \url{https://github.com/hbing-l/chronos}.}
% In this work, we investigate the limitations of current adaptation paradigms under continuous knowledge drift.

\begin{figure}[t]
    \centering
    \includegraphics[width=\linewidth]{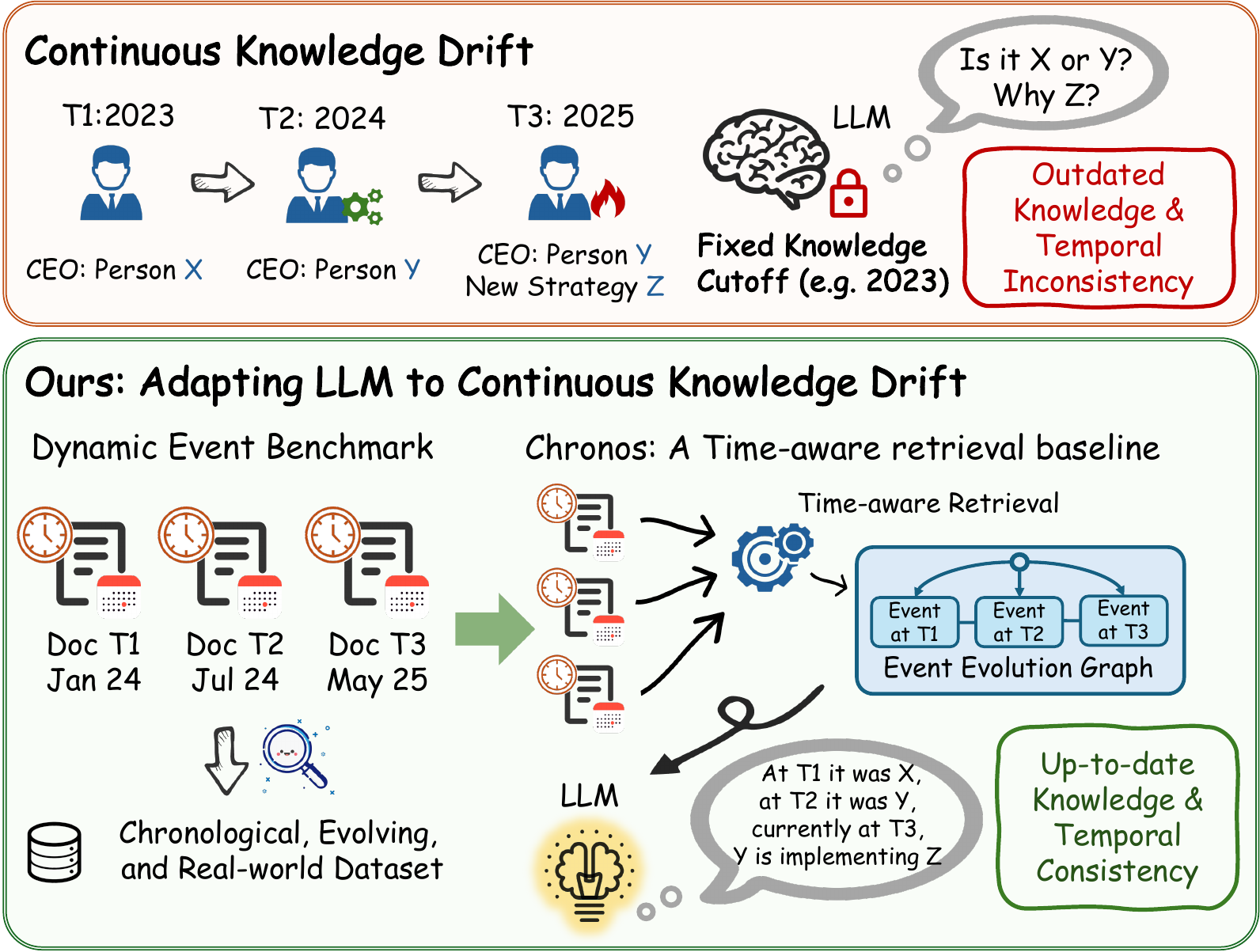}
    \caption{Illustration of continuous knowledge drift and our proposed method. Real-world knowledge evolves over time (e.g., changes in company CEO), while LLMs with a fixed knowledge cutoff may produce outdated or temporally inconsistent answers. We introduce a dynamic event benchmark composed of time-stamped documents that reflect such evolution, and a time-aware retrieval baseline (\method) that progressively organizes retrieved evidence into an Event Evolution Graph, enabling more temporally consistent reasoning.}
    \label{fig:teaser}
    \vspace{-1.1em}

\end{figure}

\section{Introduction}
% \method
% Event Evolution Graph
% % Sacred Timeline
% historical knowledge preservation
% time-aware knowledge management
% dependency-aware Knowledge Updating
% heterogeneous knowledge graph

Large language models (LLMs) have shown strong performance across a wide range of tasks, including question answering and reasoning, by acquiring extensive linguistic and factual knowledge during pretraining \cite{achiam2023gpt,he2024llm,sun2025survey,liu2025bingo}. However, this knowledge is inherently limited to a fixed snapshot of the world, commonly referred to as a knowledge cutoff \cite{chang2024large,thedewikibigedit}. In contrast, real-world knowledge evolves continuously: organizations reorganize, officials change roles, policies are revised, and entity attributes are updated over time \cite{wang2025bring}. As a result, deployed LLMs may gradually become misaligned with the current state of the world, a phenomenon we refer to as \textbf{\textit{continuous knowledge drift}}.

Beyond outdated facts, continuous knowledge drift also poses challenges for temporal reasoning \cite{chu2024timebench,wang2024tram,qiu2024large}. As knowledge evolves, correct reasoning requires models to not only access up-to-date information, but also preserve past knowledge and maintain temporal consistency in how events and facts relate and change over time. This involves capturing their ordering, dependencies, and state transitions. Without such temporal consistency, models may produce inconsistent or logically incorrect outputs even when relevant evidence is available \cite{guo2025temporal}. These challenges naturally motivate the central question of this work:

% Beyond outdated facts, continuous knowledge drift also poses challenges for temporal reasoning \cite{chu2024timebench,wang2024tram,qiu2024large}. Correct reasoning often requires not only accessing up-to-date information, but also retaining past knowledge and understanding how events evolve, relate to one another, and unfold over time. Without such temporal awareness, models may produce inconsistent or logically incorrect outputs even when relevant evidence is available. These challenges motivate the central question of this work:
% \begin{quote}
% \emph{How can LLMs adapt to continuously evolving real-world knowledge while maintaining temporally consistent understanding?}
% \end{quote}

\begin{tcolorbox}[
  % colback=gray!5,
  % colframe=gray!60,
  colback=langgreen!5,
  colframe=langgreen!60,
  boxrule=0.8pt,
  arc=6pt,
  left=6pt,
  right=6pt,
  top=6pt,
  bottom=6pt
]
\centering
\emph{How can LLMs adapt to continuously evolving real-world knowledge while maintaining temporally consistent understanding?}
\end{tcolorbox}

A natural approach to addressing knowledge drift is continual pretraining or finetuning \cite{shi2025continual,zheng2025towards,zheng2025lifelong}. While effective in some settings, such methods are computationally expensive, require repeated retraining as knowledge evolves, and are prone to catastrophic forgetting \cite{ke2023continual}. Knowledge editing offers a more targeted alternative by modifying a small subset of model parameters to update specific facts \cite{meng2022locating,mengmass,mitchell2021fast,li2024pmet,wang2024wise}. However, existing editing methods are primarily designed to overwrite a model’s current beliefs, without explicitly preserving past knowledge. Moreover, they typically operate on one fact at a time, whereas real-world knowledge updates often arrive in batches. When applied repeatedly, individual edits may interfere with each other, and models lack a principled mechanism to distinguish past from current information.

Retrieval-augmented generation (RAG) provides a complementary paradigm by incorporating external knowledge at inference time without modifying model parameters, and has shown strong performance on a wide range of knowledge-intensive tasks \cite{fan2024survey,yu2024rankrag}. However, existing RAG approaches are rarely evaluated in settings that reflect chronological, evolving, and real-world knowledge evolution. Most evaluations focus on isolated or static updates, and seldom require reasoning over sequences of evolving events. As a result, it remains unclear whether retrieval-based methods can support temporally consistent understanding under continuous real-world knowledge drift. Additional discussion of related work is provided in Appendix~\ref{ap:related_work}.

Addressing this gap requires evaluation settings that reflect how knowledge evolves over time. However, existing benchmarks are largely static or centered on single-fact updates, and seldom require handling ordered event sequences or enforcing temporal consistency \cite{thedewikibigedit,levy2017zero,jang2022temporalwiki}. In addition, many commonly used benchmarks rely on knowledge that is already outdated by one or more years. To bridge this gap, we introduce a new benchmark of real-world dynamic events from 2024 to 2025, spanning ten domains. The benchmark collects time-stamped evidence describing both newly emerging and historically changing facts, thereby reflecting real-world continuous knowledge drift and enabling the study of how models adapt to evolving knowledge over time.

Building on this benchmark, we further introduce \textbf{\method}, a time-aware retrieval framework designed for continuously evolving knowledge, as illustrated in Figure~\ref{fig:teaser}. \method incorporates temporal signals into retrieval and progressively organizes evidence through an \emph{Event Evolution Graph}, which explicitly encodes how entities and events change over time. By structuring evidence in this way, \method enables more temporally consistent understanding without modifying model parameters.

Overall, our contributions are threefold:
\begin{enumerate}[leftmargin=*, itemsep=0pt, labelsep=5pt, topsep=3pt]
    \item \textbf{Problem formulation.} We formalize the problem of adapting LLMs to continuously evolving real-world knowledge, highlighting the challenge of maintaining temporally consistent understanding as knowledge changes over time.
    
    \item \textbf{A new benchmark.} We introduce a new benchmark of real-world dynamic events. The benchmark consists of time-stamped evidence reflecting how knowledge changes in practice, and enables systematic evaluation of models under continuous knowledge drift.
    
    \item \textbf{A strong baseline.} We propose \method, a retrieval-based framework that constructs an Event Evolution Graph from retrieved evidence to model how entities and events evolve over time, which enables more temporally consistent understanding, and serves as a strong baseline for the proposed benchmark.
\end{enumerate}

\section{Benchmark Design}
Our benchmark consists of two main components: a collection of knowledge quadruples and five categories of question–answer pairs. These question types define distinct evaluation tasks that assess models from multiple perspectives, evaluating both their ability to access up-to-date factual knowledge and to maintain temporal consistency when reasoning under continuous knowledge drift. 

\subsection{Knowledge Quadruple Curation}
We first define the temporal scope of all knowledge to span from \textit{January 1, 2024} to \textit{October 31, 2025}. Accordingly, all evaluated LLMs are assumed to have a knowledge cutoff prior to 2024, ensuring that the benchmark tests their ability to handle post-cutoff information.

% CR: October 31, 2025 -> December 31, 2025

We manually define ten topical domains, including \textit{Corporate Leadership Changes}, \textit{Sports Coaching Changes and Player Transfers}, and \textit{Economic Indicators and Policy Changes}, among others. Entities within these domains are highly dynamic, undergoing frequent updates and continuously giving rise to new events, making them particularly suitable for evaluating a model's ability to track and reason over evolving knowledge across time.

For each topical domain, we first identify a set of core entities whose attributes are prone to frequent change over time, and treat each as a subject. We then track the evolution of these entities using Wikipedia data from 2024 to 2025, recording all observable state changes and newly emerging events together with their corresponding timestamps. Each change event is extracted using large language models and represented as a knowledge quadruple of the form $(\textit{subject}, \textit{relation}, \textit{object}, \textit{timestamp})$, which explicitly captures the temporal dynamics of real-world knowledge.

% For each topical domain, we first identify a set of core entities whose attributes are prone to frequent change over time, and treat each of them as a subject. We then track the evolution of these entities during the period from 2024--2025, recording all observable state changes or new events together with their corresponding timestamps. Each change event is represented as a knowledge quadruple of the form $(\textit{subject}, \textit{relation}, \textit{object}, \textit{timestamp})$, which explicitly captures the temporal dynamics of real-world knowledge.

% Following this formulation, we track how these entities evolve according to the wikipedia data in 2024-2025, and use LLM to organizing them into temporal sequences that reflect the evolution of knowledge over time. All generated data entries are subsequently filtered, reviewed, and refined through an additional round of manual verification to ensure factual correctness.

% CR: -> Following this formulation, we manually curate several high-quality seed examples per topic. Building upon these seeds, we further employ large language models equipped with web search capabilities to generate additional candidate quadruples for each domain. All automatically generated entries are subsequently reviewed and refined through a second round of manual verification to ensure correctness and temporal consistency.

\begin{figure*}[htbp]
    \centering
    \includegraphics[width=\textwidth]{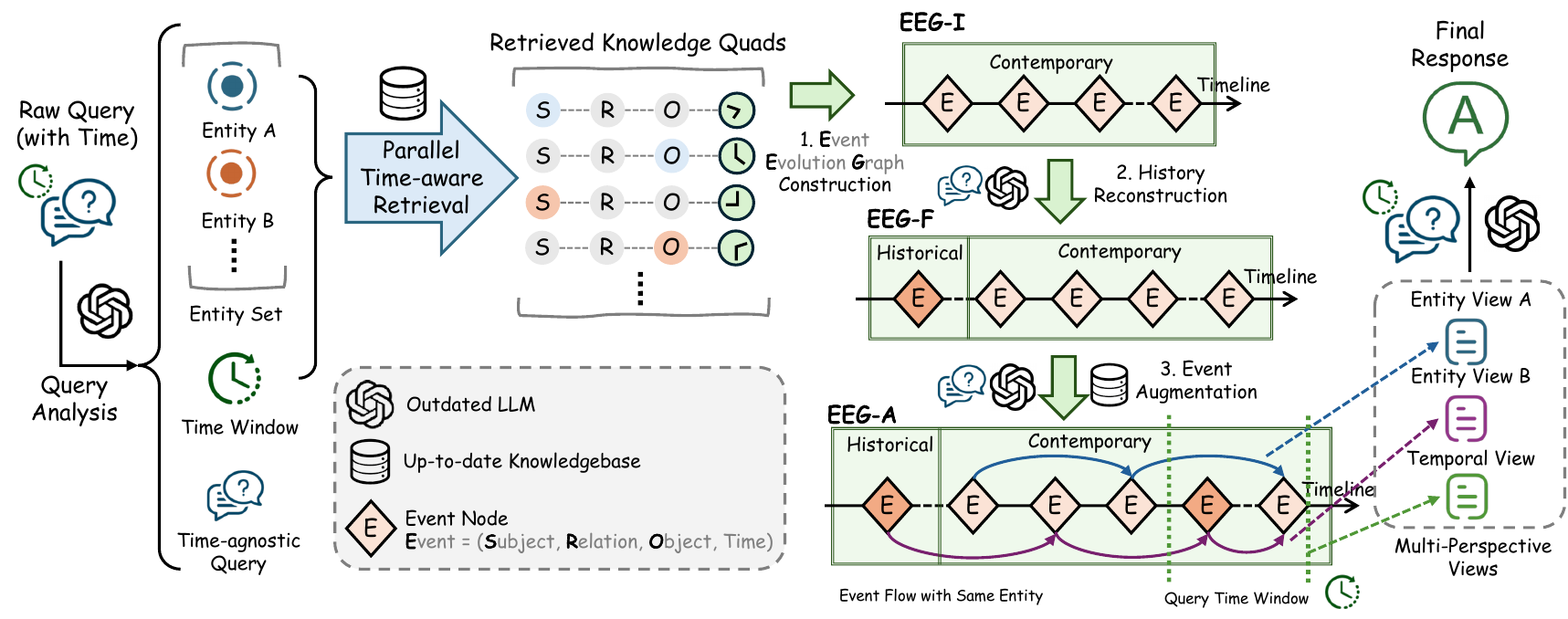}
    \caption{Overview of \method. The framework first performs query analysis to extract relevant entities and the associated time window. A time-aware retriever then collects temporally relevant knowledge quadruples from an up-to-date knowledge base. These facts are organized into an Event Evolution Graph (EEG), which models how entity states evolve over time by sorting events along a timeline and linking events that share common entities. The EEG is progressively refined, and its resulting multi-perspective views enable the model to reason over evolving knowledge and produce temporally consistent responses.}
    \label{fig:framework}
\end{figure*}

\subsection{Question-Answer Pair Construction}

\begin{table}[t]
\centering
\caption{An illustration of the dataset structure. Each instance includes time-stamped knowledge quadruples together with questions covering historical, contemporary, and commonsense settings, designed to evaluate reasoning over evolving knowledge.}
\scriptsize
\setlength{\tabcolsep}{5pt}
\renewcommand{\arraystretch}{1.25}
\begin{tabular}{p{0.94\linewidth}}
\toprule

\textit{\textbf{Knowledge}} \\

$\langle$World’s Richest Person, held by, Elon Musk, 2024-01-01$\rangle$ \\
$\langle$World’s Richest Person, held by, Bernard Arnault, 2024-01-28$\rangle$ \\
$\langle$World’s Richest Person, held by, Jeff Bezos, 2024-03-05$\rangle$ \\
$\langle$World’s Richest Person, held by, Bernard Arnault, 2024-04-02$\rangle$ \\
$\langle$World’s Richest Person, held by, Elon Musk, 2024-06-08$\rangle$ \\
$\langle$World’s Richest Person, held by, Larry Ellison, 2025-09-10$\rangle$ \\
$\langle$World’s Richest Person, held by, Elon Musk, 2025-09-11$\rangle$ \\
\cdashline{1-1}[0.5pt/1pt]
$\langle$Oracle stock price, surged to, USD 328, 2025-09-10$\rangle$ \\

\midrule

\textit{\textbf{Historical Question}} \\
Who was the richest person in the world in March 2014? \\

\midrule

\textit{\textbf{Contemporary Questions}} \\

\textit{\textbf{C1}}: Who was the world’s richest person on August 20, 2025? \\
\textit{\textbf{C2}}: Who were the world’s richest people at any point during 2024? \\
\textit{\textbf{C3}}: Which company’s stock surge led to Elon Musk losing his position as the world’s richest person on September 10, 2025? \\

\midrule

\textit{\textbf{Commonsense Question}} \\
What is the brightest star in the night sky? \\
A.~Sirius \quad
B.~Sun \quad
C.~Polaris \quad
D.~North Star \\

\midrule

\textit{\textbf{Answers}} \\
\textit{\textbf{Historical}}: Bill Gates  \quad
\textit{\textbf{Commonsense}}: A \quad
\textit{\textbf{C1}}: Elon Musk \\
\textit{\textbf{C2}}: Elon Musk, Bernard Arnault, Jeff Bezos  \quad
\textit{\textbf{C3}}: Oracle
 \\

\bottomrule
\end{tabular}
\label{tab:quadruple_example}
\end{table}

To evaluate whether models can retain historical knowledge, and understand up-to-date knowledge and correctly reason over temporally evolving information, we design five types of QA tasks that can be grouped into three major categories.

\begin{enumerate}[leftmargin=*, itemsep=2pt, labelsep=5pt, topsep=3pt]
    \item \textbf{Historical QA.}  
    This category evaluates whether a model can correctly recall historical facts that occurred before 2024. Under our setting, all information required to answer these questions originates from events prior to 2024.

    \item \textbf{Contemporary QA.}  
    This category evaluates whether a model can correctly answer questions involving up-to-date knowledge occurring between \textit{2024-01-01} and \textit{2025-10-31}, which we treat as contemporary knowledge. It is further divided into three subtypes that require different forms of temporal reasoning:
    % CR: 2025-10-31 -> 2025-12-31
    \begin{enumerate}[leftmargin=*, itemsep=2pt, labelsep=5pt, topsep=3pt]
        \item \textbf{Single-timestamp QA (C1).}  
        Each question can be answered using a single knowledge quadruple associated with one specific timestamp.

        \item \textbf{Multi-timestamp QA (C2).}  
        Each question requires reasoning over multiple temporally distinct facts associated with the same entity.

        \item \textbf{Multi-source QA (C3).}  
        Each question requires integrating information from multiple entities and multiple timestamps, testing the model’s ability to combine evidence across sources while preserving temporal consistency.
    \end{enumerate}

    \item \textbf{Commonsense QA.}  
    This category evaluates time-invariant commonsense reasoning ability and is not directly related to the other four question types or the knowledge base used in our benchmark. It serves to verify that incorporating temporally evolving knowledge does not degrade a model’s general-purpose reasoning performance.
\end{enumerate}

For QA construction, we primarily rely on large language models to generate candidate questions.  
For \textit{Historical QA}, questions are generated such that the relevant events occur strictly before 2024.  
For \textit{C1}, each question is derived directly from a single knowledge quadruple.  
For \textit{C2}, multiple quadruples associated with the same entity are merged to construct questions requiring temporal aggregation.  
For \textit{C3}, quadruples from different but semantically related entities are combined, ensuring that correct answers require multi-source reasoning across both entities and timestamps. An illustration of the dataset structure is provided in Table~\ref{tab:quadruple_example}.

All generated questions undergo manual verification to ensure clarity, correctness, and alignment with the intended evaluation setting. For \textit{Commonsense QA}, we directly adopt the TruthfulQA dataset \cite{lin2022truthfulqa} as the evaluation data.

This construction process ensures that the benchmark captures realistic knowledge evolution while maintaining high data quality and enabling comprehensive and systematic evaluation of model robustness under temporal change. Dataset statistics are provided in Appendix~\ref{ap:stat}.

\section{Methodology}

We propose \method, a time-aware retrieval-augmented framework that serves as a strong baseline for the proposed benchmark. Its overall architecture is illustrated in Figure~\ref{fig:framework}.

\subsection{Task Formulation}

Given an input query $q_{\text{raw}}$, which may contain explicit or implicit temporal constraints, we assume access to an up-to-date knowledge base $\mathcal{K}$ consisting of a set of time-stamped quadruples $\{k_i\}$, where each quadruple is defined as $k_i = (s_i, r_i, o_i, t_i)$. Here, $s_i$ denotes the subject, $r_i$ the relation, $o_i$ the object, and $t_i$ the timestamp associated with the fact. All timestamps fall within a fixed time window of newly acquired knowledge, $T_0 \le t_i \le T_1$.  

We further assume an outdated large language model $\mathcal{M}$ whose pretraining data only covers information available up to time $T_0$, and therefore lacks up-to-date knowledge $\mathcal{K}$. Under this setting, we study the problem of answering queries under continuous knowledge drift and the goal is to enable $\mathcal{M}$ to produce the correct final answer $A$ for diverse types of queries.

\subsection{Query Analysis}
Given an input query $q_\text{raw}$, the model $\mathcal{M}$ is prompted to perform query analysis:
\begin{equation}
    \mathcal{E},\; q_{0},\; t_q = \mathcal{M}(q_\text{raw} \mid \mathcal{P}_1),
\end{equation}
where $\mathcal{E}=\{e\}$ denotes the set of extracted entities, $q_{0}$ is a rewritten, time-agnostic version of the query with all temporal expressions removed, and $t_q = [t_s, t_e]$ is the time window of the query. $\mathcal{P}_1$ denotes the prompt used for this task.

It decomposes the raw query into explicit semantic components, enabling more precise and controllable retrieval. In addition, explicitly modeling temporal constraints is crucial because temporal information retrieval remains a long-standing challenge~\cite{abdallah2025extending, kanhabua2016temporal}. Dense text embeddings often struggle to accurately encode temporal expressions, making it difficult to distinguish events occurring at different points in time. By explicitly representing the time window, our approach enables time-aware retrieval that better aligns retrieved evidence with the intended temporal context of the query.

% In contrast, vanilla retrieval methods typically rely on embeddings of the raw query, which can introduce semantic ambiguity and lead to noisy or irrelevant retrieval results.

\subsection{Time-aware Retrieval}

For each extracted entity $e$, we treat it as a query $q_e$ and associate it with a time window $t_q = [t_s, t_e]$. We perform time-aware retrieval in parallel for all entities using dense semantic embeddings.

Let $\mathcal{K}_{q_e} = \{(k_i, sim_i)\}_{i=1}^{N}$ denote the top-$N$ candidate knowledge items retrieved based on semantic similarity, where $sim_i \in [0,1]$ is the similarity score between the query $q_e$ and the knowledge item $k_i$. Each knowledge item $k_i$ is associated with a timestamp $t_i$.

To incorporate temporal relevance, we define the temporal distance between the timestamp $t_i$ and the query time window $t_q$ as:
\begin{equation}
\Delta(t_i, t_q) =
\begin{cases}
t_s - t_i, & \text{if } t_i < t_s, \\
t_i - t_e, & \text{if } t_i > t_e, \\
0, & \text{if } t_s \le t_i \le t_e .
\end{cases}
\end{equation}

This definition assigns zero distance to events occurring within the target time interval and penalizes events outside the interval proportionally to their temporal deviation.

We then convert this distance into a temporal relevance score using an exponential decay function:
\begin{equation}
\mathrm{TimeScore}(k_i) = \exp\!\left(-\frac{\Delta(t_i, t_q)}{\tau}\right),
\end{equation}
where $\tau$ is a decay hyperparameter controlling how rapidly relevance decreases as the timestamp moves away from the query interval.

Finally, we compute a unified retrieval score by combining semantic similarity and temporal relevance:
\begin{equation}
\mathrm{Score}(k_i)
= \alpha \, sim_i + (1 - \alpha)\, \mathrm{TimeScore}(k_i),
\end{equation}
where $\alpha \in [0,1]$ balances the contribution of semantic similarity and temporal proximity.

All candidate knowledge items are ranked in descending order according to $\mathrm{Score}(k_i)$, and the top-$n$ items are selected to form the time-aware retrieved set:
\begin{equation}
\mathcal{K}_{r} = \text{Top-}n(\mathcal{K}_{q_e}, \mathrm{Score}).
\end{equation}

This time-aware retrieval mechanism prioritizes evidence that is both semantically relevant and temporally aligned with the query, providing a principled foundation for subsequent steps.

\subsection{Event Evolution Graph Construction}
Given a set of retrieved knowledge quadruples
$\mathcal{K}_r = \{(s_i, r_i, o_i, t_i)\}_{i=1}^{n}$, we construct an \textit{Event Evolution Graph (EEG)} to explicitly model how entities, their relations, and associated events evolve over time. The EEG is built through three progressive stages, denoted as \textit{EEG-I}, \textit{EEG-F}, and \textit{EEG-A}, corresponding to the \textbf{I}nitial, \textbf{F}ull, and \textbf{A}ugmented graphs, respectively.

\paragraph{\textit{Step 1: Initial Temporal Ordering.}}
% We first construct an initial temporal graph by ordering all retrieved events according to their timestamps. Each event is represented as a quadruple $(s_i, r_i, o_i, t_i)$ and placed along a global timeline, forming a linear structure that reflects the chronological progression of facts. Formally, we define the initial event evolution graph as
% \begin{equation}
% \mathcal{G}_{\mathrm{I}} = \mathrm{Sort}\!\left(\mathcal{K}_r,\; \text{by } t_i \right),
% \end{equation}
% where $\mathcal{K}_r$ denotes the set of retrieved events and the sorting operation arranges them in ascending order of their timestamps $t_i$. The resulting graph $\mathcal{G}_{\mathrm{I}}$, referred to as \textit{EEG-I}, captures the temporal ordering among retrieved events.

We first construct an initial temporal graph by organizing all retrieved events according to their timestamps. Each knowledge quadruple $(s_i, r_i, o_i, t_i)$ is treated as an \emph{event node} $E$ and positioned along a global timeline, forming a linear structure that reflects the chronological progression of facts. Formally, we define the initial event evolution graph as
\begin{equation}
\mathcal{G}_{\mathrm{I}} = \left(\mathcal{K}_r,\ \rightarrow_t\right),
\end{equation}
where $\mathcal{K}_r$ denotes the set of retrieved events, and $\rightarrow_t$ denotes directed temporal edges induced by their chronological ordering. Specifically, we add a temporal edge $k_i \rightarrow_t k_j$ only between two consecutive events $k_i, k_j \in \mathcal{K}_r$ such that $t_i < t_j$. The resulting graph $\mathcal{G}_{\mathrm{I}}$, referred to as \textit{EEG-I}, captures the temporal ordering among retrieved events.

\paragraph{\textit{Step 2: History Reconstruction.}}
To enable reasoning over historical context beyond retrieved facts, we further construct a more complete view of event evolution by incorporating reconstructed historical events.

Specifically, conditioned on the time-agnostic query $q_0$ and its associated time window $t_q$, we prompt the model $\mathcal{M}$ to infer historical events that are relevant to the query. This process yields a set of historical knowledge quadruples:
\begin{equation}
\mathcal{K}_{\mathcal{H}} = \mathcal{M}(q_0, t_q \mid \mathcal{P}_2),
\end{equation}
where each element in $\mathcal{K}_{\mathcal{H}}$ is a time-stamped quadruple $(s_i, r_i, o_i, t_i)$. Here, $\mathcal{P}_2$ denotes the prompt used to elicit historical knowledge.

These reconstructed quadruples are then merged with the retrieved events and inserted into the global timeline according to their inferred timestamps, yielding full graph \textit{EEG-F} $\mathcal{G}_{\text{F}}$:
\begin{equation}
\mathcal{G}_{\text{F}} = \big(\mathcal{G}_\text{I} \cup \mathcal{K}_\mathcal{H},\; \rightarrow_t \big),
\end{equation}
where $\rightarrow_t$ denotes directed temporal edges induced by chronological ordering.

\paragraph{\textit{Step 3: Event Augmentation and Entity-Centric Linking.}}
To augment the graph, we enrich it with additional relevant events. Specifically, conditioned on the raw query $q_{\text{raw}}$ and the current full graph $\mathcal{G}_{\mathrm{F}}$, we use prompt $\mathcal{P}_3$ to instruct the model $\mathcal{M}$ to either  (i) directly generate additional knowledge quadruples or (ii) produce a follow-up query $q_1$ that targets potentially missing but relevant information, which is then retrieved to obtain additional knowledge quadruples. Obtained quadruples are merged into the existing graph, forming an augmented knowledge set $\mathcal{K}_{\text{aug}}$, which improves the coverage and completeness.

% The resulting augmented event evolution graph, denoted as \textit{EEG-A}, is defined as
% \begin{equation}
% \mathcal{G}_{\text{A}} = \big(\mathcal{G}_{\mathrm{F}} \cup \mathcal{K}_{\text{aug}},\; \rightarrow_t \big),
% \end{equation}
% where temporal edges $\rightarrow_t$ encode chronological relations among all events.

While temporal ordering captures the global progression of all events, effective reasoning often requires tracking how the \textit{same entity} evolves across multiple events over time. Therefore, we further augment the graph through an entity-centric linking mechanism.

% For each entity $e$, we define the set of all events in which it participates as:
% \begin{equation}
% \mathcal{E}(e) = \{(s_i, r_i, o_i, t_i) \mid s_i = e \;\lor\; o_i = e\}.
% \end{equation}

% These events are sorted chronologically and connected via entity-specific edges:
% \begin{equation}
% e_i \rightarrow_e e_j 
% \quad \text{if} \quad 
% t_i < t_j \;\land\; e \in \{s_i, o_i\} \cap \{s_j, o_j\}.
% \end{equation}

For each entity $e$, we define the set of all events in which it appears as either subject or object:
\begin{equation}
\mathcal{K}_e = \{\, k_i \in \mathcal{K}_{\text{aug}} \mid s_i = e \ \lor\ o_i = e \,\}.
\end{equation}

% These events are sorted in ascending order of their timestamps, and consecutive events are connected by entity-specific directed edges. Formally, for any two events $k_i, k_j \in \mathcal{K}_e$, we add an entity edge:
% \begin{equation}
% k_i \;\rightarrow_e\; k_j
% \quad \text{if} \quad
% t_i < t_j.
% \end{equation}
% This construction yields a temporal chain that captures the evolution of each entity over time.

% The resulting augmented graph \textit{EEG-A} ($\mathcal{G}_{\text{A}}$) integrates three complementary structures: (a) temporal edges capturing global chronological order, (b) historical and augmented  events providing broader contextual coverage, and (c) entity-centric edges modeling how individual entities evolve across time.

These events are sorted in ascending order of their timestamps, and consecutive events are connected by entity-specific directed edges. Formally, for any two events $k_i, k_j \in \mathcal{K}_e$, we add an entity edge $k_i \rightarrow_e k_j$ if $k_j$ immediately follows $k_i$ in time for the same entity. It yields a temporal chain that captures the evolution of each entity over time.

The resulting augmented graph, denoted as \textit{EEG-A} $\mathcal{G}_{\text{A}}$, integrates three complementary components: (i) temporal edges that encode the global chronological order of events, (ii) historical and augmented events that provide broader contextual coverage, and (iii) entity-centric edges that model how individual entities evolve across time.

\subsection{Prediction with Multi-Perspective Views}

Given the augmented event evolution graph $\mathcal{G}_{\text{A}}$, we derive multiple complementary subgraph-based views to support robust reasoning. Specifically, we define two types of views:  
(i) \textbf{\textit{Temporal view.}} $\Phi_{\text{temp}}(\mathcal{G}_{\text{A}})$ denotes a subgraph that captures the global chronological structure of events within the queried time window. It is constructed by selecting events and temporal edges from $\mathcal{G}_{\text{A}}$ whose timestamps fall within the query scope, thereby representing the overall temporal progression of relevant facts. (ii) \textbf{\textit{Entity-centric views.}} For each entity $e_l$, $\Phi_{\text{ent}}^{l}(\mathcal{G}_{\text{A}})$ denotes a subgraph that captures the evolution of that entity over time. This view consists of all events involving entity $e_l$, together with the entity-specific temporal edges that link these events in chronological order.

% \begin{itemize}[leftmargin=*, itemsep=2pt, labelsep=5pt, topsep=3pt]
%     \item \textbf{\textit{Temporal view.}} $\Phi_{\text{temp}}(\mathcal{G}_{\text{A}})$ denotes a subgraph that captures the global chronological structure of events within the queried time window. It is constructed by selecting events and temporal edges from $\mathcal{G}_{\text{A}}$ according to their timestamps, thereby representing the overall temporal progression.

%     \item \textbf{\textit{Entity-centric views.}} For each entity $e_l$, $\Phi_{\text{ent}}^{l}(\mathcal{G}_{\text{A}})$ denotes a subgraph that captures the evolution of that entity over time. This view consists of all events involving entity $e_l$, together with the entity-specific temporal edges that link these events in chronological order.
% \end{itemize}

Each view is serialized into text and used as contextual evidence for reasoning. The final prediction is obtained by conditioning the model $\mathcal{}$ on the input query together with these multi-perspective representations:
\begin{equation}
A = \mathcal{M}\Bigl(
q_\text{raw},\;
\Phi_{\text{temp}}(\mathcal{G}_{\text{A}}),\;
\{\Phi_{\text{ent}}^{(l)}(\mathcal{G}_{\text{A}})\}_{i=1}^{L}
\;\mid\; \mathcal{P}_4
\Bigr),
\end{equation}
where $A$ denotes the predicted answer and $\mathcal{P}_4$ denotes the prompt used for inference.

% Given the final augmented graph $\mathcal{G}_{\text{A}}$, we derive multiple complementary views to support robust reasoning. Each view is a subgraph. Specifically, we have:
% (i) a \textit{temporal view}, which represents the global chronological sequence of events within the queried time window; and  
% (ii) a set of \textit{entity-centric views}, each describing the evolution of a specific entity across time. Each view is serialized into text as contextual evidence.

% The final prediction is obtained by conditioning on both the input query and these multi-perspective representations of EEG:
% \begin{equation}
% A = \mathcal{M}\bigl(q,\; \Phi_{\text{temp}}(\mathcal{G}_{\text{A}}),\; \{\Phi_{\text{ent}}^{(k)}(\mathcal{G}_{\text{A}})\}_{k=1}^{K} \mid \mathcal{P}_4 \bigr),
% \end{equation}
% where $A$ denotes the predicted answer and $\mathcal{P}_4$ denotes the prompt. $\Phi_{\text{temp}}(\mathcal{G}_{\text{A}})$ represents the temporal view and $\Phi_{\text{ent}}^{(k)}(\mathcal{G}_{\text{A}})$ denotes the entity-centric view corresponding to the $k$-th entity.

By explicitly structuring both global temporal evolution and entity-specific trajectories prior to generation, this formulation enables the model to reason over evolving knowledge more reliably and to produce temporally consistent answers without any additional parameter updates.

\begin{table*}[t]
\centering
\caption{Performance comparison across historical, contemporary, and commonsense question settings. Higher values indicate better accuracy. Best results for each model are shown in \textbf{bold}, and second-best results are \underline{underlined}. \textcolor{langgreen}{Green} values indicate improvements over the strongest corresponding baseline. \method achieves consistent gains across multiple evaluation settings, while other baselines exhibit varying limitations.}
\label{tab:main_results}
\setlength{\tabcolsep}{10pt}
\resizebox{\textwidth}{!}{
\small
\begin{tabular}{lcccccc}
\toprule
\multirow{2}{*}{\textbf{Method}} 
& \multirow{2}{*}{\textbf{Historical}} 
& \multicolumn{3}{c}{\textbf{Contemporary}} 
& \multirow{2}{*}{\textbf{Commonsense}} 
& \multirow{2}{*}{\textbf{Overall}} \\
\cmidrule(lr){3-5}
&  & \textbf{C1} & \textbf{C2} & \textbf{C3} &  &  \\
\midrule

\multicolumn{7}{l}{\textbf{\textit{Direct Generation}}} \\
\quad LLaMA-3.1 
& 45.95 & 11.31 & 1.38 & 1.69 & 45.53 & 21.17 \\
\quad Claude 
& \underline{55.86} & 10.53 & 2.75 & 3.39 & 62.55 & 27.02 \\
\quad GPT-4o 
& \underline{61.26} & 14.81 & 3.21 & 5.08 & 67.32 & 30.34 \\

\midrule
\multicolumn{7}{l}{\textbf{\textit{Parametric Updating Methods}}} \\
\quad ROME$_{\textit{\tiny LLaMA-3.1}}$
& 10.81 & 2.92 & 0.92 & 0.00 & 44.92 & 11.91 \\
\quad MEMIT$_{\textit{\tiny LLaMA-3.1}}$
& 25.23 & 7.99 & 1.38 & 0.00 & 42.59 & 15.44 \\
\quad WISE$_{\textit{\tiny LLaMA-3.1}}$
&  \underline{46.85} & 11.11 & 1.83 & 1.69 & 46.51 & 21.60 \\
\quad LoRA FT$_{\textit{\tiny LLaMA-3.1}}$
& 35.14 & 71.15 & 4.13 & 10.17 & 40.27 & 32.17 \\

\midrule
\multicolumn{7}{l}{\textbf{\textit{Retrieval-based Methods}}} \\
\quad Vanilla RAG$_{\textit{\tiny LLaMA-3.1}}$
& 10.81 & \underline{80.31} & 39.91 & 16.95 & 45.41 & 38.68 \\
\quad Vanilla RAG$_{\textit{Claude}}$
& 27.03 & 84.02 & 44.50 & 23.73 & 63.53 & 48.56 \\
\quad Vanilla RAG$_{\textit{\tiny GPT-4o}}$
& 9.91 & 86.74 & 50.46 & 25.42 & 64.87 & 47.48 \\
\quad ReAct RAG$_{\textit{\tiny LLaMA-3.1}}$
& 6.31 & 64.52 & \underline{47.25} &  \underline{32.20} &  \underline{47.98} &  \underline{39.65} \\
\quad ReAct RAG$_{\textit{\tiny Claude}}$
& 18.92 & \underline{91.62} & \underline{59.63} &  \underline{47.46} &  \underline{67.07} &  \underline{56.94} \\
\quad ReAct RAG$_{\textit{\tiny GPT-4o}}$
& 7.21 & \underline{93.37} & \underline{56.42} &  \underline{45.76} &  \underline{69.89} &  \underline{54.53} \\

\midrule
\multicolumn{7}{l}{\textbf{\textit{Chronos (Ours)}}} \\

\rowcolor{lightgray}
\quad Chronos$_{\textit{\tiny LLaMA-3.1}}$
& \textbf{50.45}\gain{3.6}
& \textbf{92.40}\gain{12.1}
& \textbf{63.76}\gain{16.5}
& \textbf{61.02}\gain{28.8}
& \textbf{50.55}\gain{2.6}
& \textbf{63.64}\gain{24.0} \\

\rowcolor{lightgray}
\quad Chronos$_{\textit{\tiny Claude}}$
& \textbf{57.66}\gain{1.8}
& \textbf{93.76}\gain{2.1}
& \textbf{66.97}\gain{7.3}
& \textbf{74.58}\gain{27.1}
& \textbf{72.34}\gain{5.3}
& \textbf{73.06}\gain{16.1} \\

\rowcolor{lightgray}
\quad Chronos$_{\textit{\tiny GPT-4o}}$
& \textbf{63.96}\gain{2.7}
& \textbf{96.30}\gain{2.9}
& \textbf{66.51}\gain{10.1}
& \textbf{71.19}\gain{25.4}
& \textbf{71.36}\gain{1.5}
& \textbf{73.86}\gain{19.3} \\

\bottomrule
\end{tabular}
}
\end{table*}

\section{Experiments}

\subsection{Setup}
We evaluate several representative methods commonly used to address knowledge drift, together with the new baseline introduced in this work. We consider three large language models (LLMs) covering both proprietary and open-source settings: \textit{GPT-4o} (\texttt{gpt-4o-2024-11-20}), \textit{Claude} (\texttt{claude-3-haiku-20240307}), and \textit{LLaMA-3.1} (\texttt{Meta-Llama-3.1-8B-Instruct}). These models span different architectures, training scales, and accessibility levels.

% CR: gpt-4o-2024-11-20 -> gpt-4o-mini-2024-07-18

Their respective knowledge cutoff dates are October 2023 (\textit{GPT-4o}), August 2023 (\textit{Claude}), and December 2023 (\textit{LLaMA-3.1}), all preceding our benchmark time range starting from January 2024. This setup ensures that none of the models has direct access to the evaluated knowledge during pretraining. 
% For each query, all methods are provided access to the same external knowledge base according to their respective designs, except for direct generation, to ensure a fair comparison. 
Answers are evaluated using exact-match accuracy for all QA tasks. More details of the experimental setup can be found in Appendix~\ref{ap:setup_detail}.

\subsection{Baselines}

We evaluate the following baselines. \textit{Direct generation} assesses pretrained models without any external knowledge updates. \textit{Parametric methods} include representative knowledge editing and fine-tuning approaches, such as ROME \cite{meng2022locating}, MEMIT \cite{mengmass}, WISE \cite{wang2024wise}, and LoRA-based fine-tuning \cite{hu2022lora}, which update model parameters to incorporate new information. All of these are implemented on \textit{LLaMA-3.1}. \textit{Retrieval-based methods} augment generation with external knowledge at inference time, including vanilla RAG \cite{lewis2020retrieval} and ReAct-style RAG \cite{yao2022react}, which incorporates self-reflective reasoning to iteratively guide retrieval. These methods are instantiated with different backbone models.

% These baselines enable a comprehensive comparison across static, parametric, and retrieval-based paradigms, highlighting their respective strengths and limitations under historical, contemporary, and commonsense evaluation settings.

\subsection{Main Results}

Table~\ref{tab:main_results} reports the performance of different adaptation paradigms under historical, contemporary, and commonsense evaluation settings. Several key observations can be drawn from the results.

Direct generation by LLMs fails to cope with continuous knowledge drift, as such models lack access to up-to-date information. Their reasonable performance on historical and commonsense questions reflects intact general reasoning ability and thus provides a baseline for assessing the impact of newly emerging knowledge in other methods.

\begin{tcolorbox}[
  colback=langgreen!10,
  colframe=langgreen,
  boxrule=1pt,
  arc=6pt,
  left=3pt,
  right=3pt,
  top=4pt,
  bottom=4pt,
  title=\textbf{Takeaway \#1: Parametric updating exhibits limited scalability under evolving knowledge.},
  coltitle=white,
  fonttitle=\bfseries,
  colbacktitle=langgreen!90
]
Parametric updating methods exhibit limited robustness under continuous knowledge drift, with some showing clear degradation in basic reasoning ability, suggesting catastrophic forgetting.
\end{tcolorbox}

Among these methods, LoRA fine-tuning shows slightly better performance, while other parametric approaches fail to deliver meaningful improvements. However, even LoRA fine-tuning still harms performance on historical and commonsense questions. Although prior work has shown that knowledge editing and fine-tuning can yield improvements in isolated settings, their effectiveness remains limited when adapting to continuous knowledge drift.

These results suggest that parameter-level updates struggle to preserve previously acquired knowledge while incorporating new information. Moreover, because parametric methods typically update the model either one instance at a time or in a batch-wise manner, newly acquired knowledge can overwrite both the model's original old knowledge and earlier updates, leading to additional forgetting over time. An error case analysis and further discussion of knowledge editing methods are provided in Appendix~\ref{ap:error_ke}.

\begin{tcolorbox}[
  colback=langgreen!10,
  colframe=langgreen,
  boxrule=1pt,
  arc=6pt,
  left=3pt,
  right=3pt,
  top=4pt,
  bottom=4pt,
  title=\textbf{Takeaway \#2: Traditional RAG Lacks Flexibility in Complex Temporal Reasoning.},
  coltitle=white,
  fonttitle=\bfseries,
  colbacktitle=langgreen!90
]
Vanilla RAG and ReAct-style RAG show substantial improvements on contemporary queries, but their performance degrades on historical questions. Their effectiveness also remains limited in complex temporal reasoning settings.
\end{tcolorbox}

The poor performance in C2 and C3 can be attributed to the need to aggregate multiple pieces of evidence. These settings require reasoning over information from multiple timestamps or entities and often involve ordering events across time. A single query is insufficient to capture such complexity. Even though ReAct-style RAG can retrieve additional evidence through iterative querying, the resulting retrieved content is often noisy and unstructured, making it difficult for the model to form a coherent temporal understanding. While ReAct-style RAG performs well on commonsense queries due to its multi-step reasoning process, this advantage does not directly translate to complex temporal reasoning scenarios.

\begin{tcolorbox}[
  colback=langgreen!10,
  colframe=langgreen,
  boxrule=1pt,
  arc=6pt,
  left=3pt,
  right=3pt,
  top=4pt,
  bottom=4pt,
  title=\textbf{Takeaway \#3: Explicitly modeling event evolution enables more reliable reasoning under continuous knowledge drift.},
  coltitle=white,
  fonttitle=\bfseries,
  colbacktitle=langgreen!90
]
\method consistently delivers strong and stable performance across settings and backbone models, particularly in scenarios that require aggregating and ordering multiple time-stamped facts, while maintaining performance on historical knowledge and commonsense reasoning.
\end{tcolorbox}

This suggests that explicitly organizing retrieved evidence into an \emph{Event Evolution Graph} (EEG), rather than merely retrieving relevant documents, helps preserve previously acquired knowledge while supporting temporally coherent reasoning. By structuring evidence along its temporal evolution and serializing it into multi-perspective views, the model reduces interference between old and new information, leading to more stable and reliable behavior under continuous knowledge drift. A case study is shown in Appendix~\ref{ap:case_study}.

\begin{table}[t]
\centering
\caption{Ablation study of \method. Each value reports accuracy, followed by its change relative to the full model. \textcolor{langgreen}{Green} values indicate performance improvements, while \textcolor{red}{red} values indicate performance drops.}
\label{tab:ablation}
\resizebox{\linewidth}{!}{
\begin{tabular}{lccccc}
\toprule
\textbf{Method} 
& \textbf{Hist.} 
& \multicolumn{3}{c}{\textbf{Contemporary}} 
& \textbf{Comm.} \\
\cmidrule(lr){3-5}
 &  & C1 & C2 & C3 &  \\

\midrule
\rowcolor{lightgray}
\textbf{Chronos (Ours)} 
& \textbf{63.96}
& \textbf{96.30}
& \textbf{66.51} 
& \textbf{71.19} 
& \textbf{71.36} \\

\midrule
\textit{- Time-aware Retr.}
& 62.16\loss{1.8}
& 92.01\loss{4.3}
& 57.80\loss{8.7}
& 59.32\loss{11.9}
& 71.76\gain{0.4} \\

\textit{- History Recon.}
& 6.31\loss{57.7}
& 97.66\gain{1.4}
& 67.43\gain{0.9}
& 72.89\gain{1.7}
& 71.69\gain{0.3} \\

\textit{- Event Augment.}
& 62.16\loss{1.8}
& 92.53\loss{3.8}
& 55.05\loss{11.5}
& 52.54\loss{18.7}
& 70.50\loss{0.9} \\

\textit{- Temporal View}
& 61.26\loss{2.7}
& 90.25\loss{6.1}
& 58.26\loss{8.3}
& 55.93\loss{15.3}
& 70.65\loss{0.7} \\

\textit{- Entity View}
& 60.36\loss{3.6}
& 89.86\loss{6.4}
& 49.54\loss{17.0}
& 50.85\loss{20.3}
& 71.32\loss{0.0} \\

\bottomrule
\end{tabular}
}
\end{table}

\subsection{Ablation Study}

Table~\ref{tab:ablation} presents an ablation study demonstrating that all components contribute to the overall performance improvements of \method. The variations observed on commonsense QA are negligible.

Time-aware retrieval helps retrieve more temporally relevant evidence, enabling more precise extraction of information within the target time window. While historical reconstruction may introduce some interference when answering about contemporary queries, this effect is expected and difficult to completely avoid. However, historical reconstruction plays a crucial role in improving performance on historical questions. Event augmentation further enriches the EEG by supplementing additional relevant knowledge, leading to more informative representations. Moreover, the temporal-view and entity-view components provide two different perspectives for serializing the EEG, jointly offering the model a more structured and informative context. Both views are necessary and mutually complementary for effective reasoning.

These components help the method progressively construct and refine the EEG, enabling more robust and temporally consistent reasoning under continuous knowledge drift.

\section{Conclusion}
This paper studies the limitations of current adaptation paradigms for LLMs under continuous knowledge drift. We introduce a new benchmark of real-world dynamic events that captures how knowledge evolves over time and enables systematic evaluation of temporal consistency in model reasoning. Our results show that existing approaches struggle to maintain coherent understanding as knowledge evolves. To address this, we propose a time-aware retrieval baseline based on an Event Evolution Graph, enabling temporally consistent reasoning without modifying model parameters. We believe this work provides a useful foundation for future research on adapting LLMs to continuously evolving real-world knowledge.

% eight pages

% \clearpage

\section*{Limitations}
To focus on recent and realistic knowledge changes, our benchmark is constructed under the assumption that the underlying language models have a knowledge cutoff before 2024, while the evaluation data cover up-to-date knowledge from 2024 to 2025. In real-world applications, however, the temporal scope of relevant knowledge may vary across domains and deployment settings, and should be adjusted accordingly. Moreover, practical industrial scenarios often involve more complex and heterogeneous knowledge dynamics than those captured in our benchmark, calling for richer and more challenging test cases in future work.

\section*{Acknowledgments}
This work is supported by the Natural Science Foundation of China (Grant 62371270).

\bibliography{references}

\clearpage
% \twocolumn[\DoToC]
\DoToC
% \clearpage

\appendix

\section{Ethical Considerations}
This work focuses on benchmarking and modeling the ability of large language models to reason over continuously evolving real-world knowledge. All data used in our benchmark are derived from publicly available sources and do not contain private, sensitive, or personally identifiable information. The benchmark is constructed to evaluate temporal reasoning and knowledge consistency rather than to profile individuals or make consequential decisions.

While our framework aims to improve temporal consistency under knowledge drift, it does not guarantee complete correctness, and outdated or inaccurate outputs may still occur, especially in high-stakes settings. Accordingly, systems based on this approach should be deployed with appropriate human oversight. 

Overall, we view this work as a methodological step toward better understanding and evaluating temporal adaptation in large language models, rather than as a standalone solution for real-world deployment.

\section{Related Work}
\label{ap:related_work}

\noindent\textbf{Continual Learning of LLMs.}
Prior work has explored adapting language models to evolving knowledge through continual learning, where model parameters are updated as new data becomes available \cite{shi2025continual,zheng2025towards,zheng2025lifelong,hu2022lora,liu2025reinforced}. Continual pretraining is one strategy for adapting LLMs to new domains or data streams, but it has been shown to incur substantial computational cost and can lead to catastrophic forgetting of previously learned information~\cite{ke2023continual,li2024examining}. Empirical studies further indicate that sequential fine-tuning and continual updates are prone to degrade performance on previously learned knowledge~\cite{luo2025empirical,li2024revisiting,liu2024enhancing}, highlighting the challenge of retaining historical knowledge alongside new information.

More recently, knowledge editing methods have been proposed as a lightweight alternative that directly modifies a small subset of model parameters to update specific facts~\cite{de2021editing,jiang2024learning,meng2022locating,mengmass,mitchell2021fast,li2024pmet}. Representative locate-then-edit approaches such as ROME and MEMIT identify and modify weights corresponding to targeted facts for efficient single or batched updates~\cite{meng2022locating,mengmass}, while other variants such as PMET offer subject-centric editing with precise control over localized changes~\cite{li2024pmet}. Extensions toward sequential or lifelong editing, such as WISE, introduce dual memory schemes and knowledge sharding to support repeated edits while balancing reliability, generalization, and locality~\cite{wang2024wise}. While effective for isolated edits, existing editing methods primarily focus on overwriting a model’s current beliefs rather than representing multiple historical states of knowledge. As a result, repeated edits may interfere with one another, and edited models typically lack mechanisms to retain or reason about past information alongside newly updated facts.
\\

\noindent\textbf{Retrieval-augmented Generation.}
Retrieval-augmented generation (RAG) has emerged as an effective approach for enabling large language models to access external information beyond their fixed pretraining data \cite{lewis2020retrieval,jiang2025deepretrieval,jiang2025ras,jimenez2024hipporag}, thereby mitigating issues related to outdated or incomplete knowledge without retraining the model itself~\cite{fan2024survey,yu2024rankrag,dong2025rag,yao2022react,lewis2020retrieval}. Recent work has explored mechanisms to adapt RAG systems more effectively to evolving information. For example, Adaptive-rag proposes strategies that adapt retrieval and generation to varying query complexities, enabling dynamic adjustment between iterative and single-step retrieval processes~\cite{jeong2024adaptive}. HoH introduce a dynamic benchmark highlighting how current RAG approaches struggle to handle outdated information in retrieval and generation, demonstrating the need for improved update handling~\cite{ouyang2025hoh}. On the modeling front, variants such as OG-RAG ground retrievals using ontological structures to better capture domain knowledge and improve contextual accuracy~\cite{sharma2025og}. Other work, such as MemoRAG, incorporates long-term memory into the RAG pipeline to support improved retrieval and generation performance, particularly in complex tasks where conventional RAG may fall short~\cite{qian2024memorag}. Despite these advances, most of these approaches assume a largely static retrieval corpus and do not explicitly model the temporal evolution of knowledge, limiting their ability to adapt to continuously drifting real-world information, which motivates our temporal retrieval framework.

\section{Statistics of the Benchmark Dataset}
\label{ap:stat}
The ten topical domains used during dataset construction are as follows:
\begin{enumerate}[leftmargin=*, itemsep=0pt, labelsep=5pt, topsep=3pt]
    \item \textit{Politics \& Government Leadership}
    \item \textit{Corporate Leadership \& Executive Roles}
    \item \textit{Sports, Coaching Changes \& Player Transfers}
    \item \textit{Regulation, Compliance, \& Financial Penalties}
    \item \textit{Mergers, Acquisitions, \& Strategic Partnerships}
    \item \textit{Product Releases \& Version Updates}
    \item \textit{Natural Disasters \& Climate Events}
    \item \textit{Public Health Events \& Disease Outbreaks}
    \item \textit{Scientific Discoveries \& Research Publications}
    \item \textit{Economic Indicators \& Policy Changes}
\end{enumerate}

After construction, each topical domain contains between 10 and 20 core subjects, and each subject is associated with approximately 3 to 7 temporally ordered events. In total, the benchmark comprises 513 knowledge quadruples. It includes 111 historical questions and 817 commonsense questions adopted from TruthfulQA~\cite{lin2022truthfulqa}. For contemporary questions, we construct three subcategories: 513 single-timestamp questions (C1), 218 multi-timestamp questions (C2), and 59 multi-source questions (C3), each designed to evaluate increasingly complex forms of temporal reasoning.

% CR: xxx,0 -> xxx => all divide 10

\section{Details of Experiment Setup}
\label{ap:setup_detail}
For generation, we set the temperature of all LLMs to 0 to ensure stable and deterministic outputs, while keeping all other hyperparameters at their default values. All prompts used in the experiments are provided in Appendix~\ref{ap:prompt}.

For dense retrieval, we use \texttt{all-mpnet-base-v2} as a lightweight retriever across all retrieval-based methods, retrieving the top 4 documents for each query. The retrieval pipeline is implemented using LangChain\footnote{\url{https://www.langchain.com}}. For the ReAct baseline, the maximum number of reasoning steps is set to 3. We use the cosine similarity between the query and target document embeddings as the similarity score.

For knowledge-editing baselines, we perform edits on the target objects according to their corresponding time-stamped knowledge quadruples, with edits conditioned on temporal information. Specifically, for ROME, MEMIT, and WISE, we adopt the default hyperparameters reported in their original papers. For the LoRA-based fine-tuning baseline, we use a learning rate of $1\times10^{-4}$, a batch size of 8, and rank 8, selecting the best-performing checkpoint at epoch 5. All experiments are conducted on a single NVIDIA A800 80GB GPU.

For the Commonsense QA task, we adopt the multiple-choice questions (MC1-target) from TruthfulQA \cite{lin2022truthfulqa}, using the V1 version as the evaluation dataset. All candidate answer options are explicitly provided in the prompt, and the model is instructed to select one option, allowing the responses to be evaluated using exact-match accuracy.

For our proposed baseline \method, we also retrieve the top 4 documents for each query. In the time-aware retrieval module, we set $\alpha = 0.75$ and $\tau_{\text{days}} = 180$, which achieve the best performance in our validation experiments. We observe that the time-aware retrieval mechanism is not highly sensitive to these hyperparameters and can be tuned easily and intuitively within a reasonable range.

\begin{table}[t]
\centering
% \scriptsize
% \setlength{\tabcolsep}{5pt}
% \renewcommand{\arraystretch}{1.2}
\caption{An error case of parameter-based knowledge editing under continuous knowledge drift. While WISE produces the same answer for all time-specific queries due to overwriting effects, \method correctly returns different answers conditioned on the queried timestamp, demonstrating its ability to model temporal variation in evolving knowledge.}
\resizebox{\linewidth}{!}{
\begin{tabular}{lcc}
\toprule

\textbf{Query (C1)}
& \textbf{WISE Output}
& \textbf{Chronos Output} \\

\midrule

Who was the world’s richest person on Jan 1, 2024?
& Elon Musk~$\checkmark$
& Elon Musk~$\checkmark$ \\

Who was the world’s richest person on Jan 28, 2024?
& Elon Musk~$\times$
& Bernard Arnault~$\checkmark$ \\

Who was the world’s richest person on Mar 5, 2024?
& Elon Musk~$\times$
& Jeff Bezos~$\checkmark$ \\

Who was the world’s richest person on Apr 2, 2024?
& Elon Musk~$\times$
& Bernard Arnault~$\checkmark$ \\

Who was the world’s richest person on Jun 8, 2024?
& Elon Musk~$\checkmark$
& Elon Musk~$\checkmark$ \\

\bottomrule
\end{tabular}
}
\label{tab:wise_error}
\end{table}

\section{Error Case Analysis of Knowledge Editing}
\label{ap:error_ke}

To better understand the limitations of parameter-based knowledge editing under continuous knowledge drift, we first highlight two fundamental challenges inherent to this paradigm. First, knowledge editing methods typically treat updated facts as static, rather than modeling them as temporally evolving entities. As a result, they lack mechanisms to represent multiple coexisting states of the same fact across different time points. Second, existing methods are designed to perform isolated edits and do not scale well to scenarios where many related facts must be updated jointly. When updates arrive sequentially, repeated parameter modifications can interfere with one another, leading to overwriting effects and the loss of previously edited knowledge.

To illustrate these limitations, we analyze a representative failure case of WISE \cite{wang2024wise}. As shown in Table~\ref{tab:wise_error}, the underlying knowledge concerns the identity of the world’s richest person at different time points. Although the dataset contains multiple time-stamped facts, WISE applies edits sequentially and updates model parameters in place. Consequently, later edits overwrite earlier ones, and the model collapses multiple temporal states into a single dominant answer. When queried with different time-specific questions (C1), the edited model repeatedly produces the same output corresponding to the most recent update, regardless of the queried timestamp.

This example directly reflects the two limitations outlined above. Because parameter-level editing does not explicitly model time, it cannot represent multiple coexisting states of the same fact, causing temporally distinct knowledge to collapse into a single outcome. Moreover, since edits are applied sequentially and independently, repeated updates interfere with one another, leading to overwriting effects when many related facts must be updated over time. Together, these behaviors illustrate why parameter-based editing struggles under continuous knowledge drift, and motivate the need for approaches that explicitly model temporal structure and preserve access to multiple historical states rather than encoding evolving knowledge implicitly in model parameters.

\section{Case Study}
\label{ap:case_study}
\begin{figure*}[htbp]
    \centering
    \includegraphics[width=\textwidth]{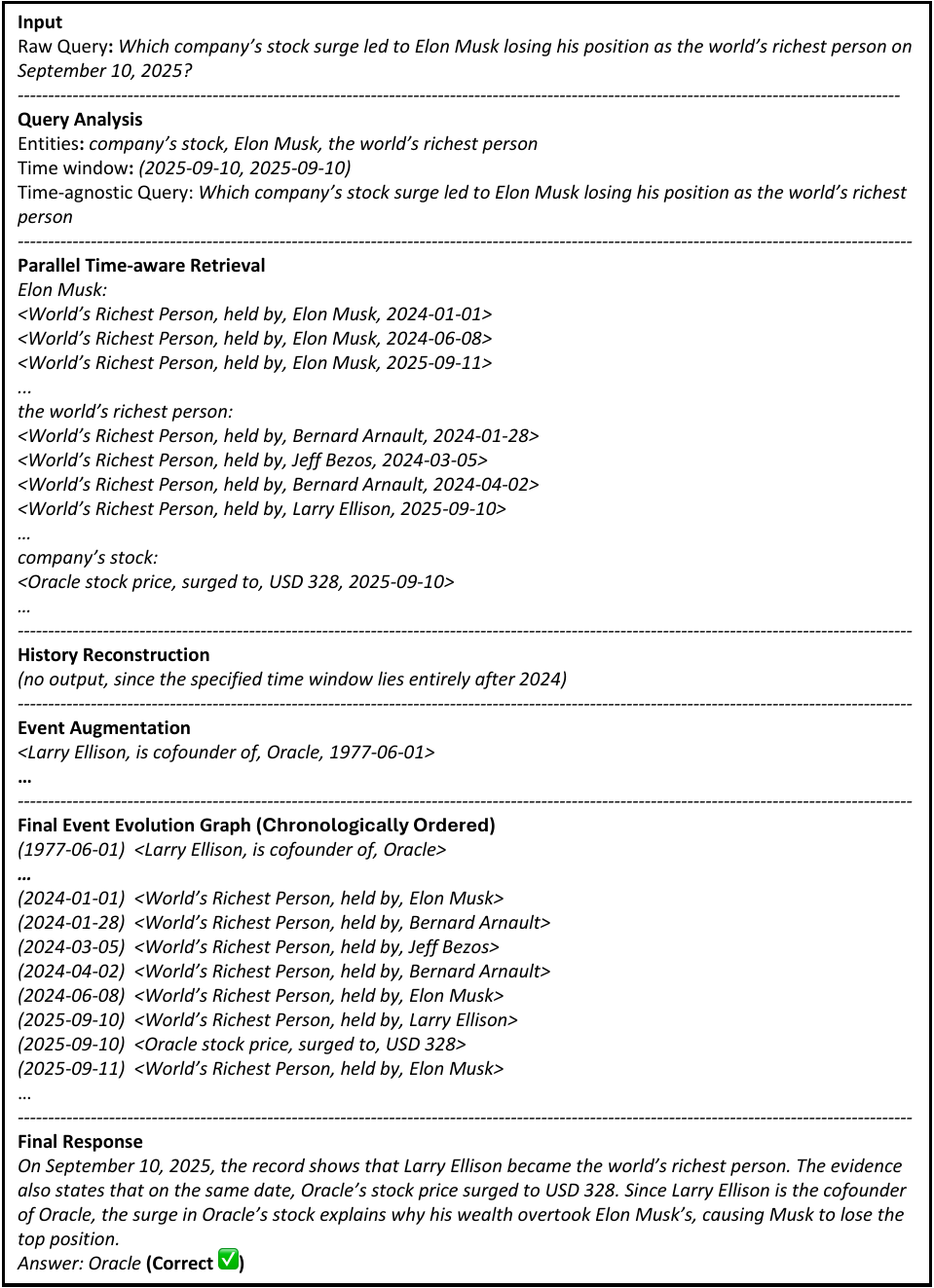}
    \caption{A case study of \method for question answering under continuous knowledge drift.}
    \label{fig:case_study}
\end{figure*}

Figure~\ref{fig:case_study} presents a case study of \method for question answering under continuous knowledge drift. The example involves a multi-source contemporary query that requires the model to jointly reason over changes in a company’s stock price and the corresponding changes in the world’s richest person. The figure illustrates the data flow through each stage of the framework, showing how evidence is progressively retrieved, augmented, and organized, ultimately leading to the correct answer.

\section{Prompt Design}
\label{ap:prompt}
\begin{figure*}[htbp]
    \centering
    \includegraphics[width=0.85\textwidth]{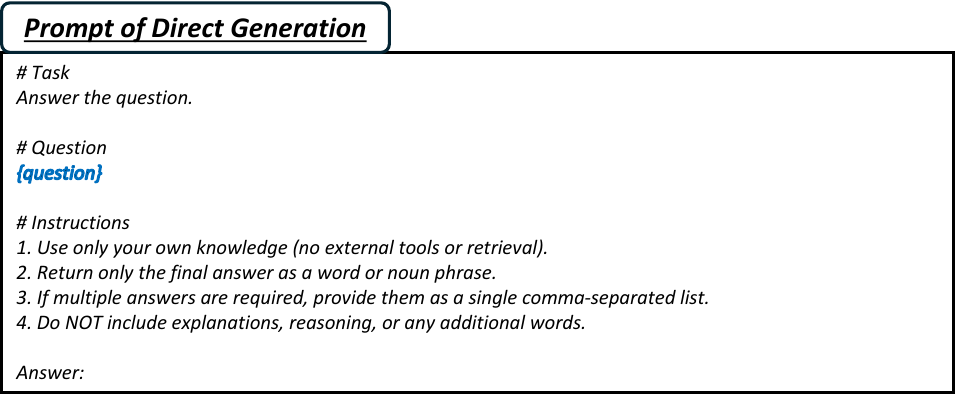}
    \caption{Prompt template used for direct generation baseline. Blue text indicates input variables.}
    \label{prompt:direct}
\end{figure*}

\begin{figure*}[htbp]
    \centering
    \includegraphics[width=0.85\textwidth]{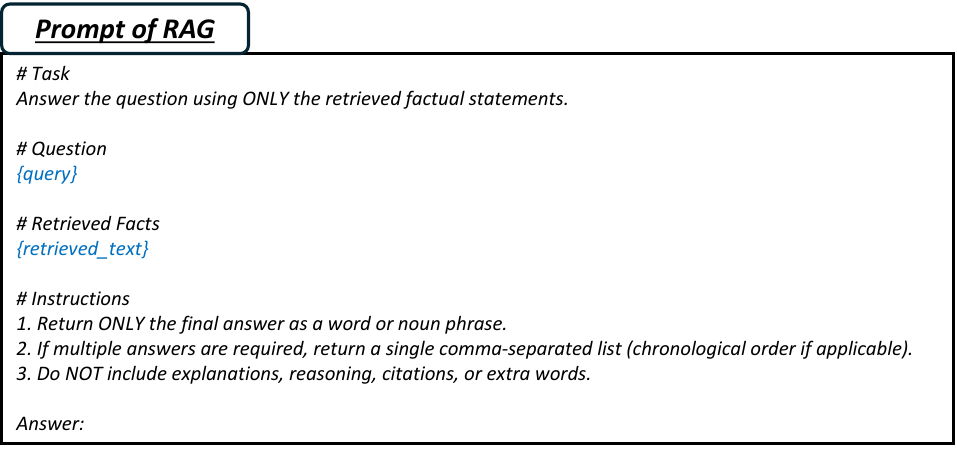}
    \caption{Prompt template used for retrieval-augmented generation baseline. Blue text indicates input variables.}
    \label{prompt:rag}
\end{figure*}

\begin{figure*}[htbp]
    \centering
    \includegraphics[width=0.85\textwidth]{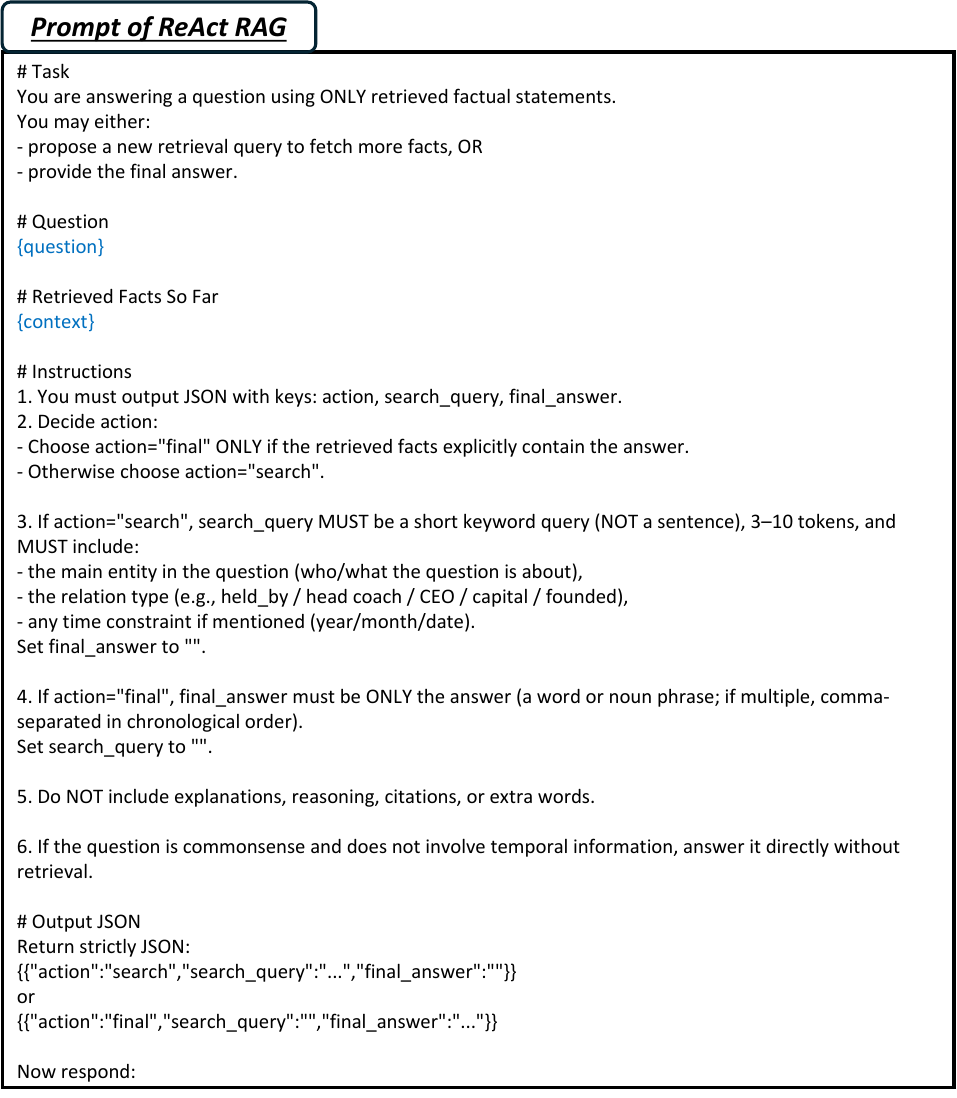}
    \caption{Prompt template used for ReAct-style retrieval-augmented generation baseline. Blue text indicates input variables.}
    \label{prompt:react}
\end{figure*}

\begin{figure*}[htbp]
    \centering
    \includegraphics[width=0.9\textwidth]{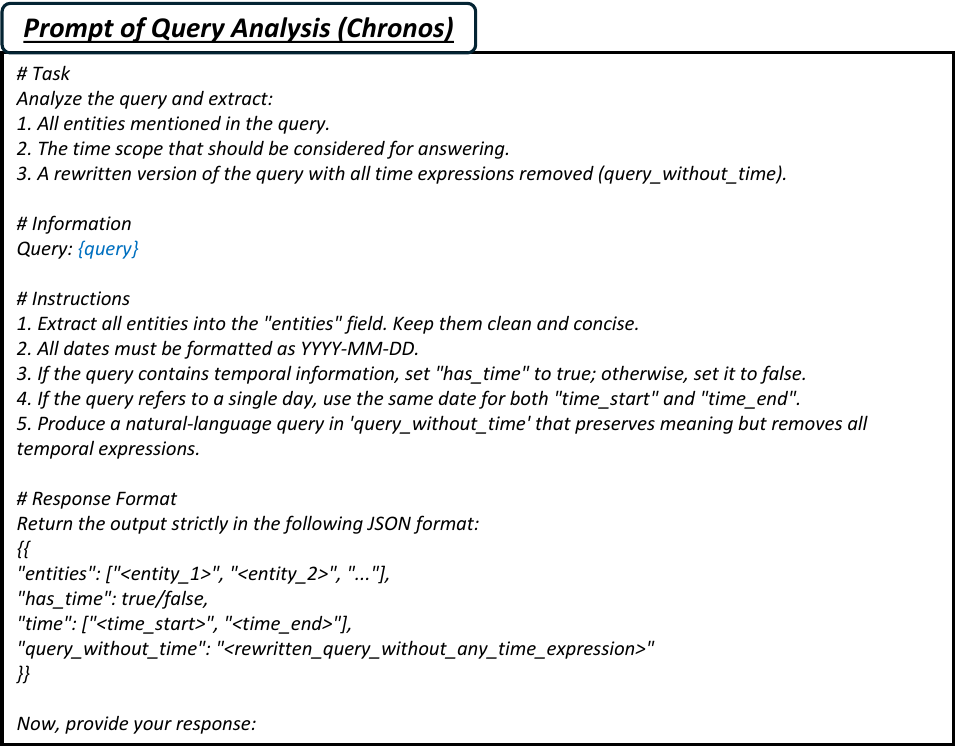}
    \caption{Prompt template used for query analysis in \method. Blue text indicates input variables.}
    \label{prompt:c_query_analysis}
\end{figure*}

\begin{figure*}[htbp]
    \centering
    \includegraphics[width=0.85\textwidth]{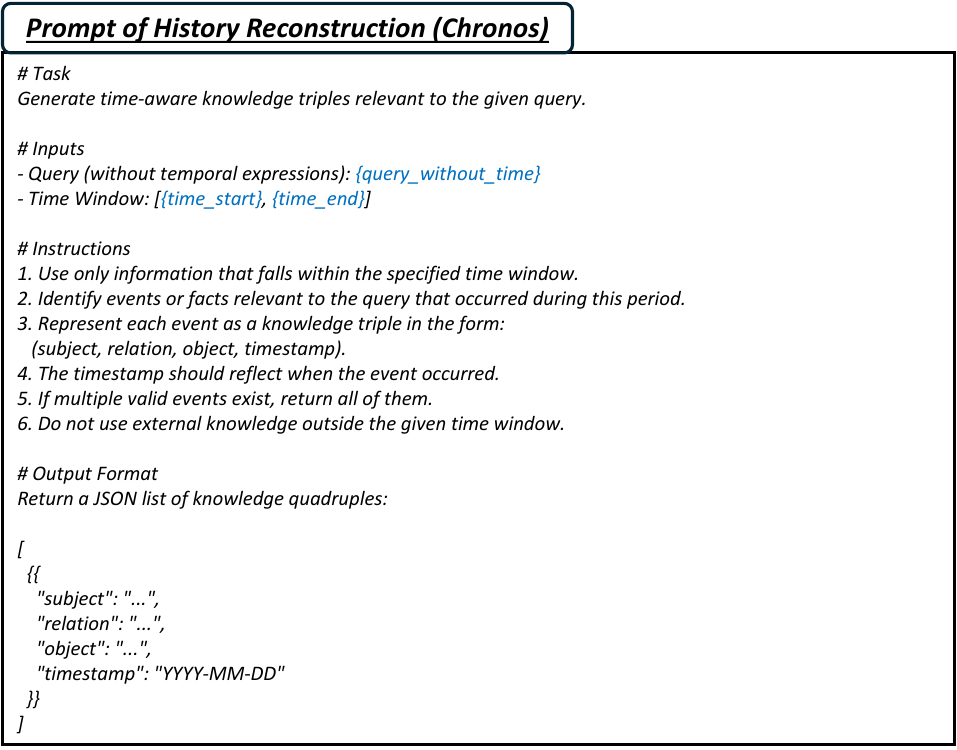}
    \caption{Prompt template used for history construction in \method. Blue text indicates input variables.}
    \label{prompt:c_history}
\end{figure*}

\begin{figure*}[htbp]
    \centering
    \includegraphics[width=0.85\textwidth]{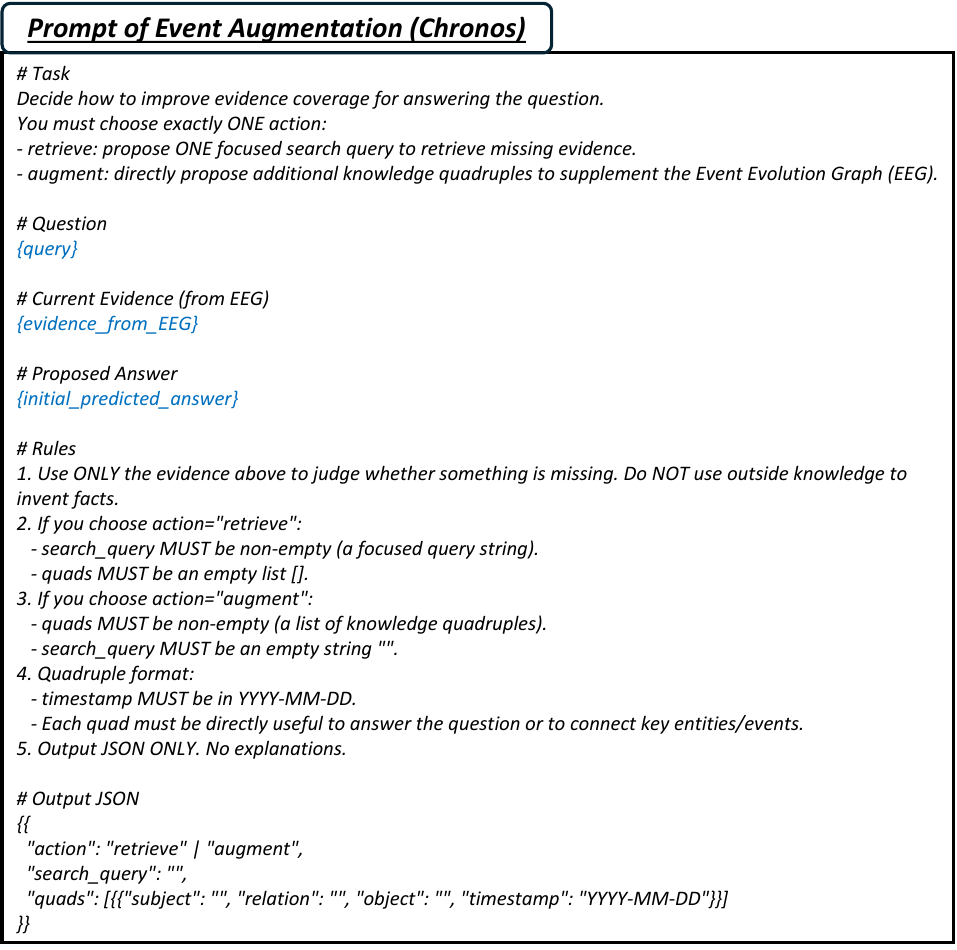}
    \caption{Prompt template used for event augmentation in \method. Blue text indicates input variables.}
    \label{prompt:c_aug}
\end{figure*}

\begin{figure*}[htbp]
    \centering
    \includegraphics[width=0.85\textwidth]{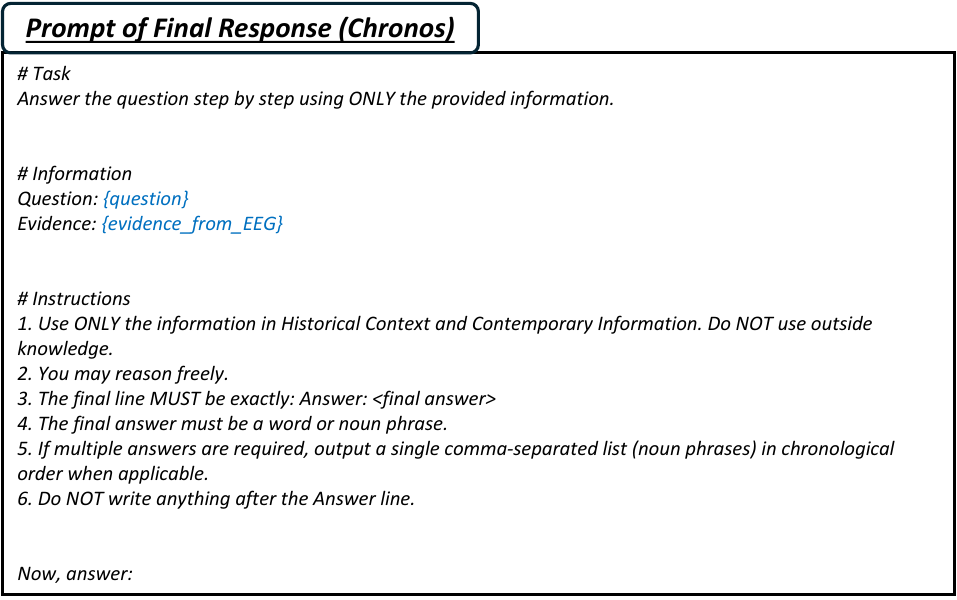}
    \caption{Prompt template used for final response in \method. Blue text indicates input variables.}
    \label{prompt:c_final}
\end{figure*}

Figures~\ref{prompt:direct}, \ref{prompt:rag}, and \ref{prompt:react} present the prompt templates adopted by the LLM-based baseline methods. Figures~\ref{prompt:c_query_analysis}, \ref{prompt:c_history}, \ref{prompt:c_aug}, and \ref{prompt:c_final} illustrate the prompts $\mathcal{P}_1$, $\mathcal{P}_2$, $\mathcal{P}_3$, and $\mathcal{P}_4$ employed at different stages of our proposed method, \method.

\end{document}